%% file: var.tex
\documentclass[10pt,twocolumn,letterpaper]{article} 

\usepackage{iccv}
\usepackage{times}

\input{preamble}

\iccvfinalcopy 
\def\iccvPaperID{3} 
\ificcvfinal\fi

\usepackage{url}

\begin{document}

\sloppy

\title{Efficient Variational Inference in Large-Scale Bayesian Compressed Sensing}

\author{George Papandreou$^1$ and Alan L\onedot Yuille$^{1,2}$\\
  $^1$Department of Statistics, University of California, Los Angeles\\
  $^2$Department of Brain and Cognitive Engineering, Korea University, Seoul, Korea\\
  \texttt{[gpapan,yuille]@stat.ucla.edu}%
}

\maketitle
\ificcvfinal
\thispagestyle{empty}
\fi

\begin{abstract}
  We study linear models under heavy-tailed priors from a probabilistic
  viewpoint. Instead of computing a single sparse most probable (MAP) solution
  as in standard deterministic approaches, the focus in the Bayesian
  compressed sensing framework shifts towards capturing the full posterior
  distribution on the latent variables, which allows quantifying the
  estimation uncertainty and learning model parameters using maximum
  likelihood. The exact posterior distribution under the sparse linear model
  is intractable and we concentrate on variational Bayesian techniques to
  approximate it. Repeatedly computing Gaussian variances turns out to be a
  key requisite and constitutes the main computational bottleneck in applying
  variational techniques in large-scale problems. We leverage on the recently
  proposed Perturb-and-MAP algorithm for drawing exact samples from Gaussian
  Markov random fields (GMRF). The main technical contribution of our paper is
  to show that estimating Gaussian variances using a relatively small number
  of such efficiently drawn random samples is much more effective than
  alternative general-purpose variance estimation techniques.
  By reducing the problem of variance estimation to standard optimization
  primitives, the resulting variational algorithms are fully scalable and
  parallelizable, allowing Bayesian computations in extremely large-scale
  problems with the same memory and time complexity requirements as
  conventional point estimation techniques. We illustrate these ideas with
  experiments in image deblurring.
\end{abstract}

\section{Introduction}
\label{sec:intro}

\paragraph{Sparsity: Deterministic and Bayesian viewpoints}

Sparsity has proven very fruitful in data analysis. Early methods such as
total variation (TV) modeling \cite{ROF92}, wavelet thresholding
\cite{Mall99}, sparse coding \cite{OlFi96}, and independent component analysis
\cite{Como94} have had big impact in signal and image modeling. Recent
research in compressed sensing \cite{CaTa05,Dono06} has shown that
high-dimensional signals representable with few non-zero coefficients in a
linear transform domain are exactly recoverable from a small number of
measurements through linear non-adaptive (typically random) operators
satisfying certain incoherence properties. Signal recovery in these
deterministic models typically reduces to a convex optimization problem and is
scalable to problems with millions of variables such as those arising in image
analysis. 

The deterministic viewpoint on sparsity has certain shortcomings. In
real-world applications the theoretical assumptions of compressed sensing are
often violated. For example, filter responses of natural images exhibit
heavy-tailed marginal histograms but are seldom exactly zero \cite{Mall89}. In
practical applications such as image inpainting or deblurring the measurement
operators are fixed and do not satisfy the incoherence properties. In these
setups it is impossible to exactly reconstruct the underlying latent signal
and it is important to quantify the associated estimation uncertainty.


Along these lines, there is a growing number of studies both in the machine
learning \cite{Atti99,LeSe00,Giro01,Tipp01,SeNi11} and the signal processing
literature \cite{JXC08,CICB10} which bring ideas from sparse modeling into a
powerful Bayesian statistical approach for describing signals and images. The
most distinctive characteristic of Bayesian modeling is that beyond finding
the most probable (MAP) solution it also allows us to represent the full
posterior distribution on the latent variables, thus capturing the uncertainty
in the recovery process. From a practical standpoint, this Bayesian compressed
sensing framework allows learning model parameters and devising adaptive
measurement designs in a principled way. We employ kurtotic priors for
modeling the heavy-tailed nature of filter responses. Beyond sparsity, we can
also capture structured statistical dependencies between model variables
\cite{Simo05} using tools from probabilistic graphical models \cite{PMK08},
yielding methods that more faithfully describe the complex statistical
properties of real-world signals.


\paragraph{Variational Bayes for sparse linear models}

Computing the exact posterior under heavy tailed priors is not tractable.  We
thus have to contend ourselves with approximate solutions, either of
stochastic sampling or deterministic variational type. In sampling techniques
we represent the posterior using random samples drawn by Markov chain
Monte-Carlo (MCMC); see \cite{PMK08,SGR10,PaYu10,SSR10} for recent related
work.

The variational techniques in which we focus in this paper approximate the
true posterior distribution with a parameterized Gaussian which allows
closed-form computations. Inference amounts to adjusting the variational
parameters to make the fit as tight as possible \cite{WaJo08}. Mostly related
to our work are \cite{Atti99,LeSe00,Giro01,SeNi11}. There exist multiple
alternative criteria to quantify the fit quality, giving rise to
approximations such as variational bounding \cite{JGJS99}, mean field or
ensemble learning, and, expectation propagation (EP) \cite{Mink01} (see
\cite{Bish06,PWKR05} for discussions about the relations among them), as well
as different iterative algorithms for optimizing each specific
criterion. These variational criteria involve some sort of integration over
the latent variables. We should contrast this with the Laplace approximation
\cite{Bish06} which is based on a second-order Taylor expansion around the MAP
point estimate and is thus inappropriate for the often non-smooth posterior
density under the sparse linear model.

All variational algorithms we study in the paper are of a double-loop nature,
requiring Gaussian variance estimation in the outer loop and sparse point
estimation in the inner loop 
\cite{SeNi11,GCLH10,SeNi11b}. The ubiquity of the Gaussian variance computation
routine is not coincidental.  Variational approximations try to capture
uncertainty in the intractable posterior distribution along the directions of
sparsity. These are naturally encoded in the covariance matrix of the proxy
Gaussian variational approximation. Marginal Gaussian variance computation is
also required in automatic relevance determination 
algorithms for sparse Bayesian learning \cite{Mack92a} and relevance vector
machine training \cite{Tipp01}; the methods we develop could also be applied
in that context.

\paragraph{Variance computation: Lanczos vs. proposed Monte-Carlo algorithm}

Estimating Gaussian variances is currently the main computational bottleneck
and hinders the wider adoption of variational Bayesian techniques in
large-scale problems with thousands or millions of variables such as those
arising in image analysis, in which explicitly storing or manipulating the
full covariance matrix is in general infeasible. Computing variances in
Gaussian Markov random fields (GMRFs) with loops is challenging and a host of
sophisticated techniques have been developed for this purpose, which often
only apply to restricted classes of models \cite{WeFr01,MJCW08}. A
general-purpose variance computation technique \cite{PaSa82,ScWi01} is based
on the Lanczos iterative method for solving eigenproblems \cite{GoVa96} and
has been extensively studied in the variational Bayes context by Seeger and
Nickisch \cite{SNPS08,SeNi11}. Unless run for a prohibitively large
number of iterations, the Lanczos algorithm severely underestimates the
required variances, to the extent that Lanczos is inadequate for optimizing
criteria like expectation propagation which are sensitive to gross variance
estimation errors \cite{SeNi11b}.

The main technical contribution of our work is to demonstrate that the
sample-based Monte-Carlo Gaussian variance estimator of \cite{PaYu10} performs
markedly better than the Lanczos algorithm as the key computational
sub-routine in the variational learning context. Our estimator builds on the
efficient Perturb-and-MAP sampling algorithm of \cite{PaYu10} (\cf
\cite{PMK08,SGR10}) which draws exact GMRF samples by locally injecting noise
to each Gaussian factor independently, followed by computing the mean/mode of
the perturbed GMRF by preconditioned conjugate gradients. Being unbiased, the
proposed sample estimator does not suffer from the Lanczos systematic
underestimation errors. In practice, a few samples suffice for capturing the
variances with accuracy sufficient for even the more sensitive expectation
propagation algorithm to work reliably. Moreover, correlations (\ie
off-diagonal elements of covariance matrix) needed in certain applications are
easy to compute.

The advocated approach to Monte-Carlo variance estimation for variational
learning has several other advantages. It is fully scalable, only relying on
well-studied computational primitives, thus allowing Bayesian inference with
the same memory and time complexity requirements as conventional point
estimation. The proposed algorithm is parallelizable, since the required
Gaussian samples can be drawn independently on different processors. Further,
we show how we can use the samples to estimate the free energy and monitor
convergence of the algorithm at no extra cost.

\section{Variational Bayes for sparse linear models}
\label{sec:glm}

\subsection{The sparse linear model: Point estimation vs. Bayesian inference}

The formulation of the sparse linear model we consider follows the setup of
\cite{Giro01,SeNi11}. We consider a hidden vector $\xv \in \Real^N$ which
follows a heavy-tailed prior distribution $P(\xv)$ and noisy linear
measurements $\yv \in \Real^M$ of it are drawn with Gaussian likelihood
$P(\yv|\xv)$. Specifically:
\begin{equation}
  \label{eq:glm}
    P(\xv;\thetav) \propto \prod_{k=1}^K t_k(\gv_k^T \xv)\mcomma \quad
    P(\yv|\xv;\thetav) = \normal{\yv}{\Hm \xv}{\sigma^2\Unitm}\mcomma
\end{equation}
where the $K$ rows of $\Gm = [\gv_1^T;\dots;\gv_K^T]$ and the $M$ rows of $\Hm
= [\hv_1^T;\dots;\hv_M^T]$ are two sets of length-$N$ linear filters, the
former mapping $\xv$ to the domain $\sv = \Gm \xv$ in which it exhibits sparse
responses and the latter capturing the Gaussian measurement
process\footnote{$\normal{\xv}{\muv}{\Sigmam}=\abs{2\pi\Sigmam}^{-1/2}\exp
  \left( -\tfrac{1}{2} (\xv-\muv)^T \Sigmam^{-1} (\xv-\muv) \right)$ is the
  multivariate Gaussian density on $\xv$ with mean $\muv$ and covariance
  $\Sigmam$.}. The sparsity inducing potentials are denoted by $t_k(s_k)$. The
Laplacian $t_k(s_k) = e^{-\tau_k \abs{s_k}}$, $s_k = \gv_k^T \xv$, is a widely
used form for them. In some applications a subset of the model's aspects
$(\Hm, \sigma^2, \Gm)$ can be unknown and dependent on a parameter vector
$\thetav$; \eg, in blind image deconvolution $\thetav$ typically is the
unknown blurring kernel $\kv$ which determines the measurement matrix $\Hm$.

By Bayes' rule, the posterior distribution of the latent variables $\xv$ given
$\yv$ has the non-Gaussian density
\begin{align}
  \label{eq:x-cond-y}
  P(\xv|\yv) = Z^{-1}(\thetav) P(\yv|\xv) \prod_{k=1}^K t_k(s_k)\mcomma
  \quad \textrm{where}\\
  \label{eq:y}
  Z(\thetav) \triangleq P(\yv;\thetav)=\int P(\yv|\xv) \prod_{k=1}^K t_k(s_k) \,d\xv
\end{align}
is the evidence/ partition function. 

Point estimation corresponding to standard compressed sensing amounts to
finding the posterior MAP configuration $\hat{\xv}_{\textrm{MAP}} \triangleq
\argmax_\xv \log P(\xv|\yv)$, leading to minimization of
\begin{equation}
  \label{eq:map}
  \phi_{\textrm{MAP}} (\xv) = \sigma^{-2} \norm{\yv-\Hm\xv}^2 -2\sum_{k=1}^K \log t_k(s_k) \mdot
\end{equation}
Point estimation thus reduces to a standard optimization problem and a host of
modern techniques have been developed for solving it, scalable to large-scale
applications. However, since it ignores the partition function, point
estimation neither provides information about the estimation uncertainty nor
allows parameter estimation.

In the Bayesian framework we try to overcome these shortcomings by capturing
the full posterior distribution. Since it is intractable to manipulate it
directly, we consider variational approximations of Gaussian form
\begin{multline}
  \label{eq:q-x-cond-y}
    Q(\xv|\yv) \propto P(\yv|\xv) e^{\betav^T\sv - \frac{1}{2}
      \sv^T\Gammam^{-1}\sv} = \normal{\xv}{\hat{\xv}_Q}{\Am^{-1}}\mcomma \, \textrm{with}\\
    \hat{\xv}_Q = \Am^{-1}\bv \mcomma \quad 
    \Am = \sigma^{-2}\Hm^T\Hm + \Gm^T\Gammam^{-1}\Gm \mcomma\\
    \Gammam = \diag(\gammav) \mcomma \quad \textrm{and} \quad
    \bv = \sigma^{-2} \Hm^T\yv + \Gm^T\betav \mdot
\end{multline}
The implied form for the variational evidence is
\begin{equation}
  \label{eq:q-y}
  Z_Q(\thetav) \triangleq Q(\yv;\thetav) = \int P(\yv|\xv) e^{\betav^T\sv - \frac{1}{2} \sv^T\Gammam^{-1}\sv} d\xv \mdot
\end{equation}
Our task in variational learning is to adjust the set of variational
parameters $\xiv = (\betav,\gammav)$ so as to improve the fit of the
approximating Gaussian to the true posterior distribution.

We will mostly be focusing on log-concave sparsity inducing potentials
$t_k(\cdot)$ -- \ie, $\log t_k(\cdot)$ is concave -- such as the Laplacian. This
guarantees that the posterior $P(\xv|\yv)$ is also log-concave in $\xv$, and
thus point estimation in \eq~\eqref{eq:map} is a convex optimization
problem. Log-concavity also implies that $P(\xv|\yv)$ is unimodal and
justifies approximating it with a Gaussian $Q(\xv|\yv)$ in
\eq~\eqref{eq:q-x-cond-y}.

\subsection{Variational bounding}
\label{sec:vb}

Variational bounding \cite{JGJS99,Giro01,PWKR05,SeNi11} is applicable to
sparsity-inducing potentials of super-Gaussian form. The family of even
super-Gaussian potentials is quite rich and superset of the family of mixtures
of zero-mean Gaussians; it includes the Laplacian and the Student as members
\cite{PWKR05}. Super-Gaussian potentials have a useful dual representation
\begin{align}
  \label{eq:super-gauss}
  t_k(s_k) &= \sup_{\gamma_k > 0} e^{-s_k^2/(2\gamma_k) - h_k(\gamma_k)/2} \mcomma
  \quad \textrm{with}\\
  \label{eq:super-gauss-dual}
  h_k(\gamma_k) &\triangleq \sup_{s_k} -s_k^2/\gamma_k - 2\log t_k(s_k)
\end{align}
Variational bounding amounts to replacing the potentials $t_k(s_k)$ in
\eq~\eqref{eq:x-cond-y} with these bounds and tuning the variational
parameters $\gammav$ ($\betav$ is fixed to zero in this case) so as the
variational evidence lower bounds as tightly as possible the exact evidence $Z
\ge Z_Q$. This leads to the variational free energy minimization problem (see
\cite{SeNi11} for the derivation) $\inf_{\gammav \succ \zerov}
\phi_Q(\gammav)$, where
\begin{equation}
  \label{eq:var-prob}
  \phi_Q(\gammav) = \log\abs{\Am} + h(\gammav) + \inf_\xv R(\xv,\gammav) \mcomma
\end{equation}
with $h(\gammav) \triangleq \sum_{k=1}^K h_k(\gamma_k)$ and $R(\xv,\gammav)
\triangleq \sigma^{-2} \norm{\yv-\Hm\xv}^2 + \sv^T\Gammam^{-1}\sv$. The $\Am$
and $\bv$ are given in \eq~\eqref{eq:q-x-cond-y}; note that $\Am$ is a
function of $\gammav$.

The log-determinant term in \eq~\eqref{eq:var-prob} is what makes Bayesian
variational inference more interesting and at the same time computationally
more demanding than point estimation. Indeed, using
\eq~\eqref{eq:super-gauss}, we can re-write the objective function for MAP
estimation \eqref{eq:map} as $\phi_{\textrm{MAP}} (\xv) = \inf_{\gammav \succ
 \zerov} h(\gammav) + R(\xv,\gammav)$, showing that $\phi_{\textrm{MAP}}$ and
$\phi_Q$ only differ in the $\log\abs{\Am}$ term, which endows variational
inference with the ability to capture the effect of the partition
function. The difficulty lies in the fact that the elements of the vector
$\gammav$ are interleaved in $\log\abs{\Am}$. Following \cite{PWKR05,SeNi11},
we can decouple the problem by exploiting the concavity of $\log\abs{\Am}$ as
a function of $\gammav^{-1} \triangleq
(\gamma_1^{-1},\dots,\gamma_K^{-1})$. Fenchel duality then yields the upper
bound $\log\abs{\Am} \le \zv^T \gammav^{-1} - g^*(\zv)$, $\zv \succ \zerov$. For
given $\gammav$ the bound becomes tight for $\zv = \nabla_{\gammav^{-1}}
\log\abs{\Am} = \diag (\Gm\Am^{-1}\Gm^T)$, which can be identified as the
vector of marginal variances $z_k = \Var_Q(s_k|\yv)$ along the directions $s_k
= \gv_k^T\xv$ under the variational posterior $Q(\xv|\yv)$ with the current
guess for the parameters $\gammav$.

This approach naturally suggests a \textbf{double-loop} algorithm, globally
convergent when the potentials $t_k$ are log-concave \cite{PWKR05,SeNi11}. In
the \emph{outer loop}, we compute the vector of marginal variances $\zv$ so as
to tighten the upper bound to $\log\abs{\Am}$, given the current value of
$\gammav$.

In the \emph{inner loop}, instead of $\phi_Q$ in \eq~\eqref{eq:var-prob} we
minimize \wrt $\xv$ and $\gammav$ the upper bound given the newly computed
$\zv$
\begin{multline}
  \label{eq:var-prob-decouple-xg}
  \bar{\phi}_Q(\xv,\gammav;\zv) = \zv^T \gammav^{-1} + h(\gammav) + R(\xv,\gammav)\\
  = \sigma^{-2} \norm{\yv-\Hm\xv}^2 + \sum_{k=1}^K
  \left( \frac{s_k^2+z_k}{\gamma_k} + h_k(\gamma_k) \right) \mdot
\end{multline}
We can minimize this expression explicitly \wrt $\gammav$ by noting that it is
decoupled in the $\gamma_k$ and recalling from \eqref{eq:super-gauss} that
$-2\log t_k(s_k) = \inf_{\gamma_k>0} s_k^2/\gamma_k +
h_k(\gamma_k)$. This leaves us with a minimization problem \wrt $\xv$ alone
\begin{multline}
  \label{eq:var-prob-decouple-x}
  \bar{\phi}_Q(\xv;\zv) = \inf_{\gammav \succ \zerov} \bar{\phi}_Q(\xv,\gammav;\zv) = \\
  = \sigma^{-2} \norm{\yv-\Hm\xv}^2 -2 \sum_{k=1}^K \log t_k \left(
    (s_k^2 + z_k)^{1/2} \right) \mdot
\end{multline}
This is just a smoothed version of the MAP point estimation problem
\eqref{eq:map}, also convex when $t_k$ are log-concave, which we minimize in
the course of the inner loop with standard quasi-Newton methods \cite{BLN95}
to obtain the variational mean $\hat{\xv}$. After completion of the inner
loop, we recover the minimizing values for the variational parameters
$\gamma_k^{-1} = -2 \frac{d\log t_k(\sqrt{v})}{dv}\eat{v=\hat{s}_k^2 + z_k}$,
with which we update the vector of marginal variances $\zv$ in the subsequent
outer loop iteration \cite{SeNi11}.

\subsection{Mean field and expectation propagation}
\label{sec:ep}

Bounding is not the only way to construct variational approximations to the
intractable posterior distribution $P(\xv|\yv)$. The mean field (or ensemble
learning) approach amounts to assuming a simplified parametric form $Q$ for
the posterior distribution and adjusting the corresponding variational
parameters $\xiv$ so as to minimize the KL-divergence $D_{KL}(Q||P)$ between
$Q$ and $P$ \cite{Atti99}. See \cite{LWDF11} for a recent application of the
mean field approximation to the problem of image deconvolution, where it is
shown that the mean field updates reduce to point estimation and variance
computation primitives, exactly as in the variational bounding approximation
discussed in detail in \sect~\ref{sec:vb}.

Expectation propagation (EP) is yet another powerful variational approximation
criterion, in which the variational parameters of the approximating
distribution $Q$ are adjusted so as expectations under $Q$ and the true
posterior $P(\xv|\yv)$ are matched \cite{Mink01}. There are various iterative
sequential message passing-like algorithms for optimizing the EP
criterion. Applying EP to large-scale problems in which our paper focuses is
challenging. We will employ the parallel update algorithm of \cite{GCLH10},
but our methods are also applicable to the recently proposed provably
convergent double-loop algorithm of \cite{SeNi11b}. Once more, variance
estimation in the outer loop is the computational bottleneck; see
\cite{GCLH10,SeNi11b}.

\section{Monte-Carlo posterior variance estimation}
\label{sec:variance}

As highlighted in \sect~\ref{sec:glm}, repeatedly computing posterior Gaussian
variances turns out to be a key computational routine in all variational
approximations of the sparse linear model. With reference to
\eq~\eqref{eq:q-x-cond-y}, our goal is to compute certain elements of the
covariance matrix $\Sigmap \triangleq \Am^{-1}$ or marginal variances $z_k =
\Var_Q(s_k|\yv)$ along certain projections $s_k = \gv_k^T\xv$ under the
variational posterior $Q(\xv|\yv)$. Note that $\Sigmap$ is a fully dense
$\by{N}{N}$ matrix. Thus for large-scale models comprising $N\approx 10^6$
variables it is impossible to compute or store the full $\Sigmap$ explicitly.

\subsection{Lanczos variance estimation}
So far, the main candidate for variance estimation in the context of
large-scale variational Bayes has been the Lanczos iterative method
\cite{SNPS08,SeNi11}. As the iteration progresses, the Lanczos algorithm
builds a monotonically increasing estimate for the variances \cite{GoVa96}. It
can reveal in relatively few iterations the rough structure and relative
magnitude of variances, but requires a very large number of iterations to
accurately approximate their absolute values. Since it scales badly with the
number of iterations $N_L$ (its complexity is $\Oh(N_L^2)$ in time and
$\Oh(N_L)$ in memory due to a required reorthogonalization step), it is only
practical to run Lanczos for a relatively small number of iterations, yielding
gross underestimates for the variances.

In practice, variational bounding has proven relatively robust to the Lanczos
crude variance estimates \cite{SNPS08,SeNi11}, while expectation propagation
completely fails \cite{SeNi11b}. This starkly contrasting qualitative behavior
in the two cases can be explained as follows: In the presence of Lanczos
variance underestimation errors, the expression
\eqref{eq:var-prob-decouple-xg} remains an upper bound of \eqref{eq:var-prob},
albeit not tight any more. Moreover, the variational optimization problem
\eqref{eq:var-prob-decouple-x} gracefully degrades to the point estimation
problem \eqref{eq:map} when $0 \le \hat{z}_k \ll z_k$. In other words, despite
the variance errors the algorithm does not collapse, although it effectively
ends up solving a modified inference problem rather than the one that it was
supposed to solve. In contrast, expectation propagation works by moment
matching and the gross variance estimation errors make iterative EP
algorithms hopelessly break down.

\subsection{Efficient Monte-Carlo variance estimation with Perturb-and-MAP sampling}
\label{sec:var-sample}

We propose estimating variances using a sampling-based Monte-Carlo
technique, leveraging on the efficient Perturb-and-MAP GMRF sampling algorithm
of \cite{PaYu10}. Although \cite{PaYu10} has already suggested this
possibility, it has not explored its effectiveness in the variational Bayesian
context.

The algorithm of \cite{PaYu10} reduces GMRF sampling into a GMRF mean
estimation problem. In our notation, an exact Gaussian sample $\tilde{\xv}
\sim \normals{\zerov}{\Am^{-1}}$, with $\Am = \sigma^{-2}\Hm^T\Hm +
\Gm^T\Gammam^{-1}\Gm$, can be drawn by solving the linear system
\begin{equation}
  \label{eq:pm-sample}
  \Am \tilde{\xv} = \sigma^{-2} \Hm^T\tilde{\yv} + \Gm^T\tilde{\betav} \mdot
\end{equation}
The local perturbations $\tilde{\yv} \sim \normals{\zerov}{\sigma^2 \Um}$ and
$\tilde{\betav} \sim \normals{\zerov}{\Gammam^{-1}}$ are trivial to sample
since $\Gammam$ is diagonal. We efficiently solve the linear
system~\eqref{eq:pm-sample} using preconditioned conjugate gradients (PCG)
\cite{GoVa96}, employing filtering routines for fast evaluation of
matrix-vector products $\Am \xv$, thus avoiding the costly Cholesky
factorization step typically associated with Gaussian simulation. In contrast
to Lanczos, the memory footprint of PCG is small as only 4 length-$N$ vectors
need to be stored, while multiple samples can be trivially drawn in parallel
(using, \eg, \verb+parfor+ in Matlab). Also note that, unlike conjugate
gradients, employing preconditioning within Lanczos variance estimation is
difficult \cite{ScWi01} and seldom used in practice.

Having drawn $N_s$ Gaussian samples as described, we employ the standard
sample-based covariance estimators
\begin{equation}
  \label{eq:variance-est}
  \hat{\Sigmap} = \frac{1}{N_s} \sum_{i=1}^{N_s} \tilde{\xv}_i \tilde{\xv}_i^T 
  \mcomma \quad
  \hat{z}_k = \frac{1}{N_s} \sum_{i=1}^{N_s} \tilde{s}_{k,i}^2 \mcomma
\end{equation}
with $\tilde{s}_{k,i} \triangleq \gv_k^T \tilde{\xv}_i$. The variance
estimates marginally follow scaled chi-square distributions with $N_s$ degrees
of freedom $\hat{z}_k \sim \frac{z_k}{N_s} \chi^2(N_s)$. This implies that
$\expect{\hat{z}_k}=z_k$, \ie, this estimator is unbiased, unlike the Lanczos
one. Its relative error is $r = \Delta(\hat{z}_k)/z_k =
\sqrt{\Var(\hat{z}_k)}/z_k = \sqrt{2/N_s}$, independent from the problem size
$N$. The error drops quite slowly with the number of samples ($N_s=2/r^2$
samples are required to reach a desired relative error $r$), but variance
estimates sufficiently accurate for even the more sensitive expectation
propagation algorithm to work reliably can be obtained after about 20 samples
(which translates to $r \approx 32\%$). One can show that $z_k \le
\gamma_k^{-1}$ \cite{SeNi11}, a consequence of the fact that measurements
always reduce the uncertainty in Gaussian models. To enforce this important
structural constraint, we use in place of \eqref{eq:variance-est} the clipped
estimator $\bar{z}_k = \min(\hat{z}_k,\gamma_k^{-1})$ which behaves
considerably better in practice while still being (asymptotically) unbiased.

\begin{figure}[!htbp]
  \centering
  \includegraphics[width=0.6\columnwidth]{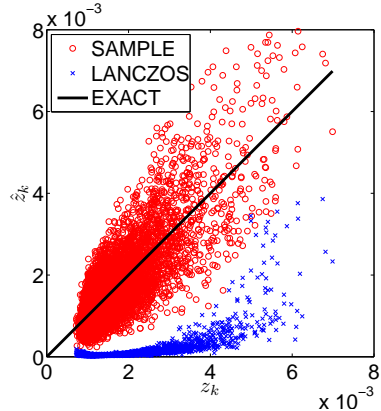}
  \caption{Scatter-plot of exact $z_k$ \vs estimated $\hat{z}_k$ marginal
    variances for a small-scale deblurring problem. We compare the the
    proposed sample-based Monte-Carlo estimator with Lanczos.}
  \label{fig:var-sample-lanczos}
\end{figure}

To illustrate the efficiency of the proposed Monte-Carlo variance estimator in
the context of variational Bayesian inference, we compare in
\fig~\ref{fig:var-sample-lanczos} the marginal variances obtained by our
sample-based estimator with that of Lanczos. The system matrix $\Am$ for this
particular example is the one of the last iteration of the double-loop
variational bounding algorithm of \sect~\ref{sec:vb} applied to a small-scale
\by{48}{73} deblurring problem for which it is feasible to compute the exact
marginal variances $z_k$. We use the clipped version $\bar{z}_k$ of our
estimator with $N_s=20$ samples, each drawn by solving the linear
system~\eqref{eq:pm-sample} with 20 PCG iterations as detailed in
\sect~\ref{sec:deblur}. Lanczos was run for $N_L=300$ iterations, so as the
runtime for the two algorithms to be the same. We see that the proposed
sample-based variance estimator performs markedly better than Lanczos, which
grossly under-estimates the true marginal variances. Note that for large-scale
problems the performance gap will be even more pronounced: as we showed
earlier, the relative estimation accuracy $r$ of the sample-based estimator is
independent of the latent space dimensionality $N$, while the relative
accuracy of Lanczos further deteriorates for large $N$ \cite[\fig~6]{SeNi11}.

\subsection{Monte-Carlo free energy estimation}

To monitor convergence of the free energy~\eqref{eq:var-prob} and for
debugging purposes, it is desirable to estimate $\log\abs{\Am}$ during the
course of the algorithm. Note that this step is not a requisite for the
variational algorithm to yield estimates for $\xv$ or estimate the model
parameters $\thetav$.

By coercing information from the samples $\tilde{\xv} \sim
\normals{\zerov}{\Am^{-1}}$ drawn for variance estimation, we can reliably
estimate $\log\abs{\Am}$ at no extra cost, provided that we can analytically
compute $\log\abs{\Pm}$ for some matrix $\Pm$ that approximates $\Am$ well,
typically the preconditioner employed by PCG for solving
\eqref{eq:pm-sample}. To see this, note that $\expect{\exp \left(
    0.5\tilde{\xv}^T(\Am-\Pm)\tilde{\xv} \right)} = \abs{\Am}/\abs{\Pm}$,
which suggests the Monte-Carlo estimator
\begin{equation}
  \label{eq:free-eng-est}
  \log\abs{\Am} \approx \log\abs{\Pm}
  - \log N_s
  + \log \left( \sum_{i=1}^{N_s} 
    0.5\tilde{\xv}_i^T(\Am-\Pm)\tilde{\xv}_i \right)
    \mdot
\end{equation}
A special case of this with $\Pm=\Um$ has been proposed before \cite{CPT05},
but for the large-scale problems we consider here using a good reference $\Pm
\approx \Am$ is crucial for the estimator \eqref{eq:free-eng-est} to exhibit
low variance and thus be useful in practice.

\section{Applications to image deconvolution}
\label{sec:deblur}

Our main motivation for this work is solving inverse problems in image
analysis and low-level vision such as image deblurring, inpainting, and
tomographic reconstruction. These give rise to large-scale inference problems
involving millions of variables. We report experimental results on image
deconvolution. Our software builds on the \verb+glm-ie+ Matlab toolbox
\cite{Nick10} designed for variational inference under the variational
bounding \cite{SeNi11} and expectation propagation \cite{GCLH10} criteria,
which we have extended to include implementations of the proposed algorithms;
our extensions will be integrated in future releases of \verb+glm-ie+.

In image deblurring \cite{Jain89,HNL06}, our goal is to recover the sharp
image $\xv$ from its blurred version $\yv$. We assume a spatially homogeneous
degradation, typically due to camera or subject motion, captured by the
measurement process $\yv = \Hm \xv \triangleq \kv * \xv$. In the non-blind
variant of the problem, the convolution blur kernel $\kv$ is considered known
(the problem is classically known as image restoration), while in the more
challenging blind variant our goal is to recover both the sharp image and the
unknown blurring kernel. 

\paragraph{Blind image deconvolution} In the blind deconvolution case, the blurring
kernel is considered as parameter, $\thetav = \kv$, which we recover by
maximum (penalized) likelihood. It is crucial to determine $\kv$ by first
integrating out the latent variables $\xv$ and then maximizing the marginal
likelihood $\argmax_{\kv} P(\yv;\kv)$, instead of maximizing the joint
likelihood $\argmax_{\kv} \left( \max_{\xv} P(\xv,\yv;\kv) \right)$
\cite{FSH+06,LWDF09}. Under the variational approximation, we use
$Q(\yv;\kv)$ from \eqref{eq:q-y} in place of $P(\yv;\kv)$. Following
\cite{Giro01,LWDF11}, we carry out the optimization iteratively using
expectation-maximization (EM).

In the \emph{E-step}, given the current estimate $\kv^t$ for the blurring
kernel, we perform variational Bayesian inference as described in
\sect~\ref{sec:glm}. In the \emph{M-step} of the $t$-th iteration, we maximize
\wrt $\kv$ the expected complete log-likelihood $\expectt{\kv^{t}}{\log
  Q(\xv,\yv;\kv)}$, with expectations taken \wrt $Q(\xv|\yv;\kv^{t})$. The
updated kernel $\kv^{t+1}$ is obtained by minimizing \wrt $\kv$ (see
\cite{LWDF11} for the derivation)
\begin{equation}
  \label{eq:m-step}
  \begin{split}
    &\expectt{\kv^{t}}{\frac{1}{2}\norm{\yv-\Hm\xv}^2}\\
    &= \frac{1}{2} \trace \left( (\Hm^T\Hm)(\Am^{-1}+\hat{\xv}\hat{\xv}^T) \right)
    -\yv^T\Hm\hat{\xv} + \mathrm{(const)}\\
    &= \frac{1}{2} \kv^T \Rm_{\xv\xv} \kv - \rv_{\xv\yv}^T \kv + \mathrm{(const)}
    \mcomma
  \end{split}
\end{equation}
which is a quadratic program in $\kv$; see \cite{LWDF11} for the formulas for
$\rv_{\xv\yv}$ and $\Rm_{\xv\xv}$. The entries in $\rv_{\xv\yv}$ accumulate
cross-correlations between $\hat{\xv}$ and $\yv$; we use the variational mean
$\hat{\xv}$ of~\eqref{eq:var-prob-decouple-x} for computing them. The entries
in $\Rm_{\xv\xv}$ capture second-order information for $\xv$ under
$Q(\xv|\yv;\kv^t)$; we estimate them efficiently by drawing a small number of
samples (1 or 2 suffice) from $\normals{\zerov}{\Am^{-1}}$, exactly as in
\sect~\ref{sec:var-sample}. Note that \cite{LWDF11} estimates $\Rm_{\xv\xv}$
by making the simplifying assumption that $\Am$ is diagonal, which could
potentially lead to a poor approximation. We add to~\eqref{eq:m-step} an extra
$L_1$ penalty term $\lambda_1\norm{\kv}_{L_1}$ so as to favor sparse kernels.

It is important to note that while the M-step update for $\kv$
in~\eqref{eq:m-step} is a convex optimization problem, the overall
log-likelihood objective $-\log Q(\yv;\kv)$ is not convex in $\kv$. This means
that the EM algorithm can get stuck to local minima. Various techniques have
been developed to mitigate this fundamental problem, such as coarse-to-fine
kernel recovery, gradient domain processing, or regularization of the result
after each kernel update with~\eqref{eq:m-step} -- see
\cite{FSH+06,LWDF09}. We have not yet incorporated these heuristics into our
blind deconvolution implementation, and thus our software may still give
unsatisfactory results when the spatial support of the unknown blurring kernel
is large.

\paragraph{Efficient circulant preconditioning}

Our sample-based variance estimator described in \sect~\ref{sec:var-sample}
requires repeatedly drawing samples $\tilde{\xv}$. For each of the samples we
solve by PCG a linear system of the form $\Am \tilde{\xv} = \tilde{\cv}$,
where $\tilde{\cv}$ is the randomly perturbed right hand side in
\eq~\eqref{eq:pm-sample}.

The system matrix $\Am = \sigma^{-2}\Hm^T\Hm + \Gm^T\Gammam^{-1}\Gm$ arising
in image deblurring is typically poorly conditioned, slowing the convergence
of plain conjugate gradients. The key to designing an effective preconditioner
for $\Am$ is to note that $\Am$ would be a stationary operator if
$\Gammam=\bar{\gamma}\Um$, \ie, the variational parameters $\gamma_k$ were
homogeneous. Following \cite{LBU12}, we select as preconditioner the
stationary approximation of the system matrix, $\Pm = \sigma^{-2}\Hm^T\Hm +
\bar{\gamma}^{-1}\Gm^T\Gm$, with $\bar{\gamma}^{-1} \triangleq (1/K)
\sum_{k=1}^K \gamma_k^{-1}$. One can prove that $\Pm$ is the stationary matrix
nearest to $\Am$ in the Frobenius norm, \ie $\Pm = \argmin_{\Xm \in
  \mathcal{C}} \norm{\Xm-\Am}$, where $\mathcal{C}$ is the set of stationary
(block-circulant with circulant blocks) matrices \cite{LBU12}. Thanks to its
stationarity, $\Pm$ is diagonalized in the Fourier domain; by employing the
2-D DFT, we can compute very efficiently expressions of the form $\Pm^{-1}\xv$
required by PCG \cite{HNL06}. Moreover, $\log\abs{\Pm}$ is also readily
computable in the Fourier domain, allowing us to use the efficient free energy
estimator~\eqref{eq:free-eng-est} for monitoring convergence. Note that the
applicability of this preconditioner extends beyond our variance estimation
setup; \eg it could be employed in conjunction with the MCMC-based deblurring
algorithm of \cite{SSR10}.

\begin{figure}[!htbp]
  \centering
  \includegraphics[width=0.55\columnwidth]{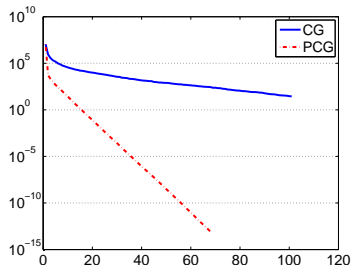}
  \caption{Conjugate gradients residual norm as function of iteration count;
    No (CG) \vs circulant (PCG) preconditioner.}
  \label{fig:pcg-res}
\end{figure}

Circulant preconditioning with $\Pm$ dramatically accelerates convergence of
conjugate gradients. We plot in \fig~\ref{fig:pcg-res} the residual in the
course of conjugate gradient iteration for a typical system matrix $\Am$
arising in deblurring a \by{190}{289} image under the variational bounding
approximation. With circulant preconditioning (PCG) we attain within only 10
iterations the same level of accuracy that is reached after 100 iterations of
unpreconditioned conjugate gradients (CG). This substantial improvement in the
convergence rate more than compensates the roughly 60\% time overhead per
iteration of PCG relative to CG (respectively, 80 \vs 50 msec per iteration on
this problem). We are not aware of any work that similarly exploits the
benefits of preconditioning in the context of Lanczos variance estimation.

\vspace{-0.5cm}

\paragraph{Image deblurring results}

We have carried out preliminary image deblurring experiments using the dataset
of \cite{LWDF09} which contains images degraded by real blur due to camera
motion, as well as their sharp versions shot with the camera still. We assume
a total-variation prior, which implies simple first-order finite difference
filters as rows of $\Gm$ and Laplacian sparsity inducing potentials $t_k(s_k)
= e^{-\tau_k \abs{s_k}}$. We fix $\tau_k=15$ which roughly matches the image
derivative scale for typical images with values between 0 and 1. We set the
noise variance to $\sigma^2=10^{-5}$.

\begin{figure*}[!htbp]
  \centering
  \begin{tabular}{c @{ } c @{ } c @{ } c}
    \includegraphics[width=0.45\columnwidth]{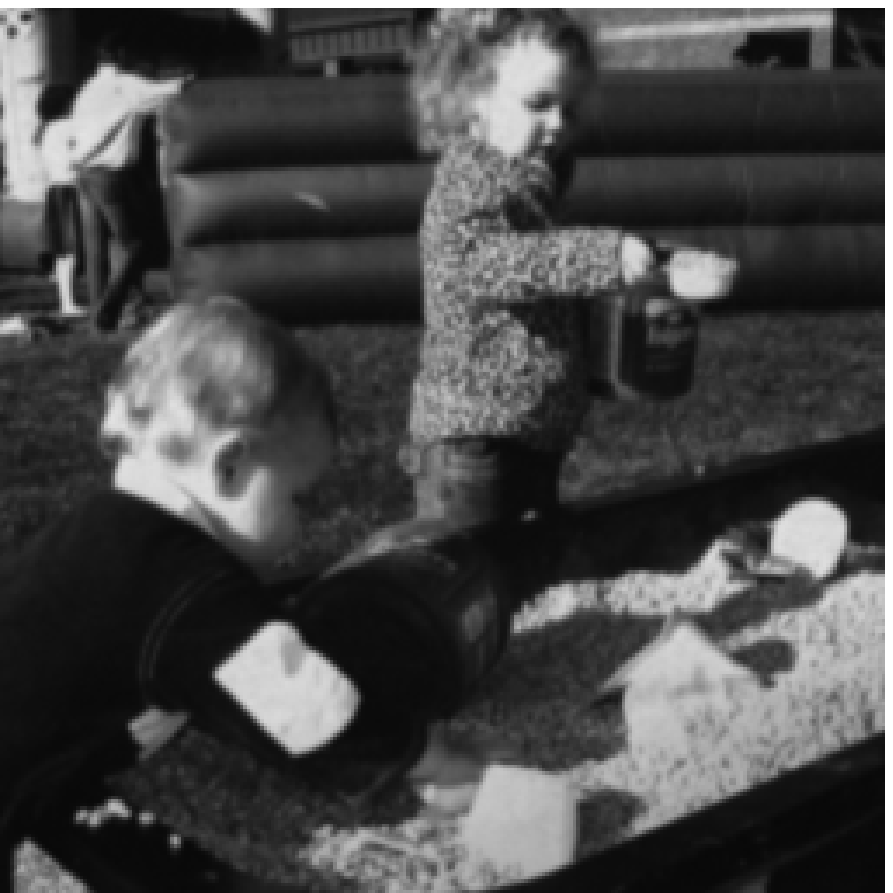}&
    \includegraphics[width=0.45\columnwidth]{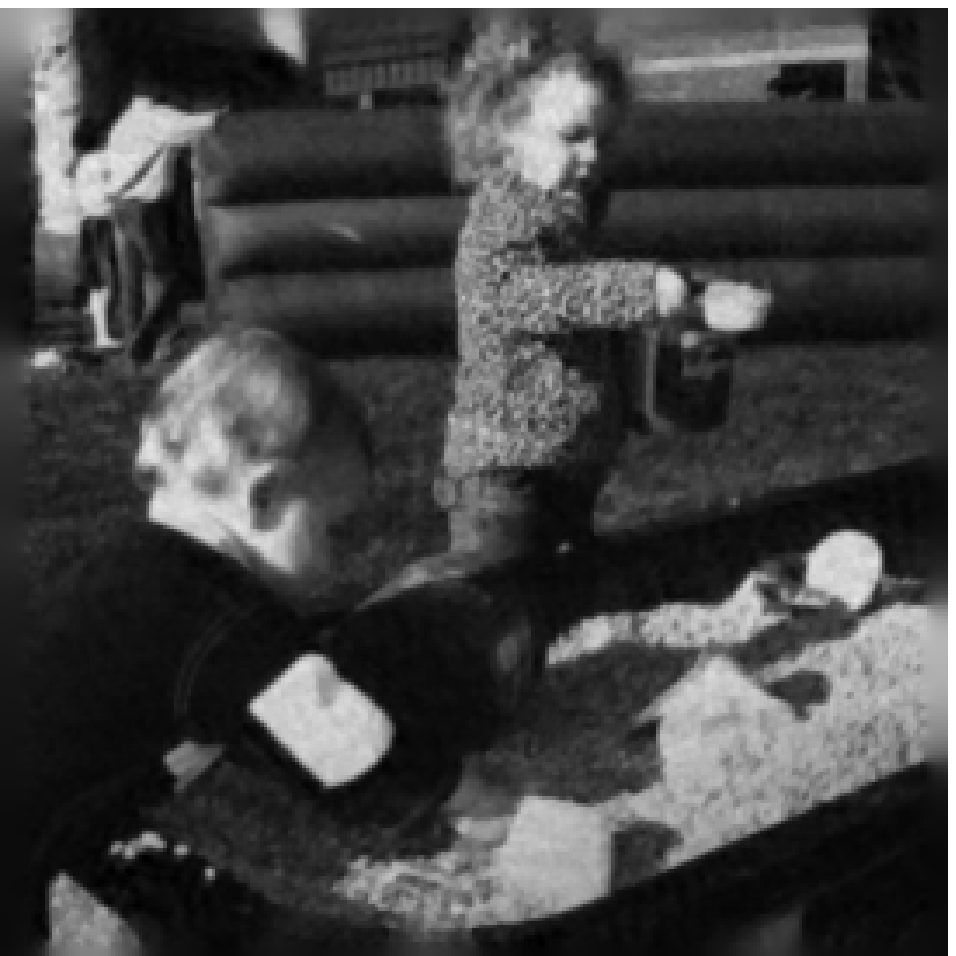}&
    \includegraphics[width=0.45\columnwidth]{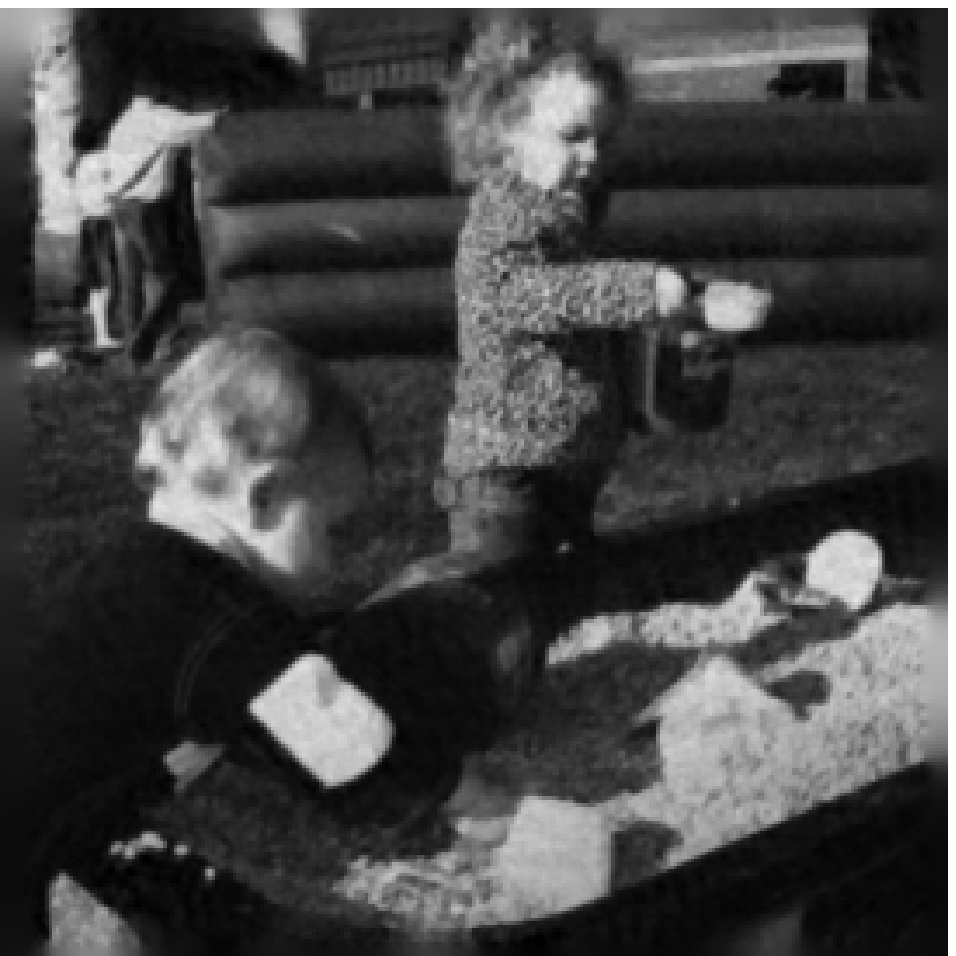}&
    \includegraphics[width=0.45\columnwidth]{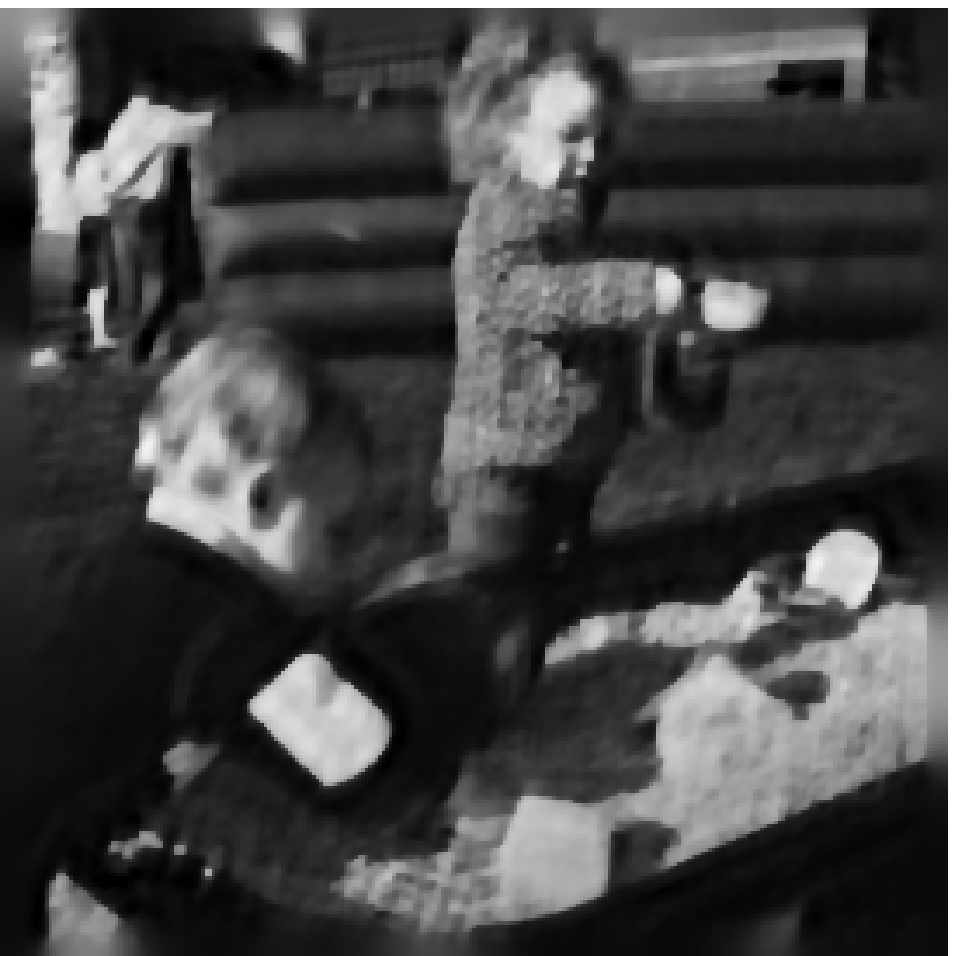}\\
    (a) sharp & (c) VB mean {\footnotesize (PSNR=31.93dB)} & 
    (e) EP mean {\footnotesize (PSNR=31.85dB)} & (g) blind VB mean {\footnotesize (PSNR=27.54dB)}\\
    \includegraphics[width=0.45\columnwidth]{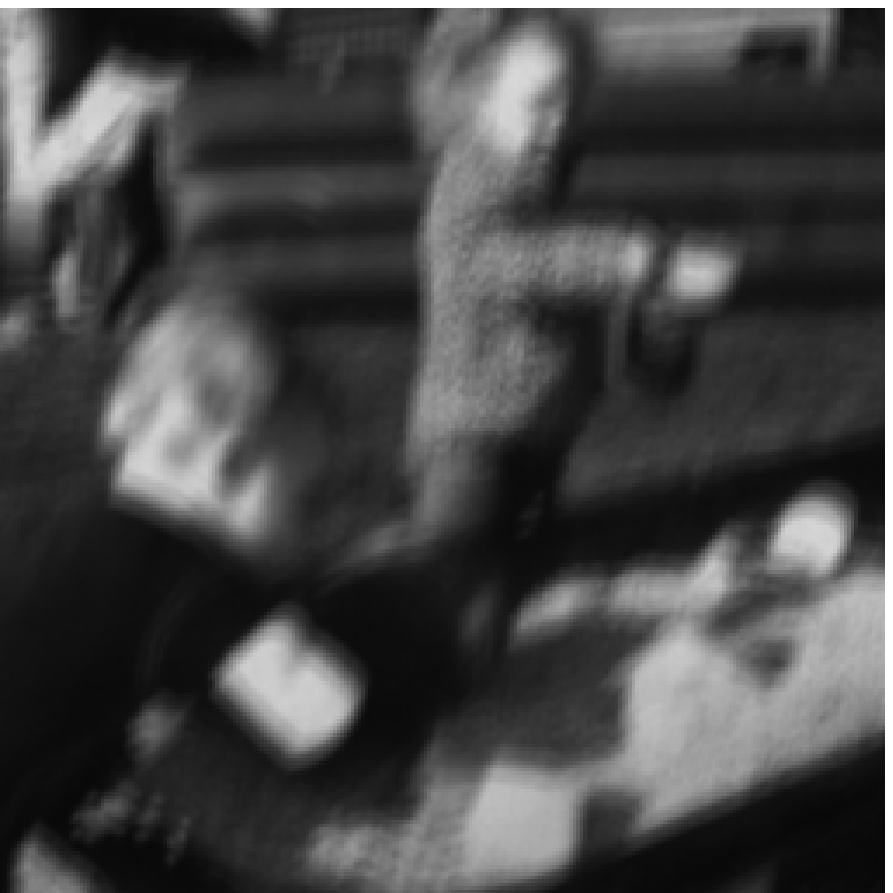}&
    \includegraphics[width=0.45\columnwidth]{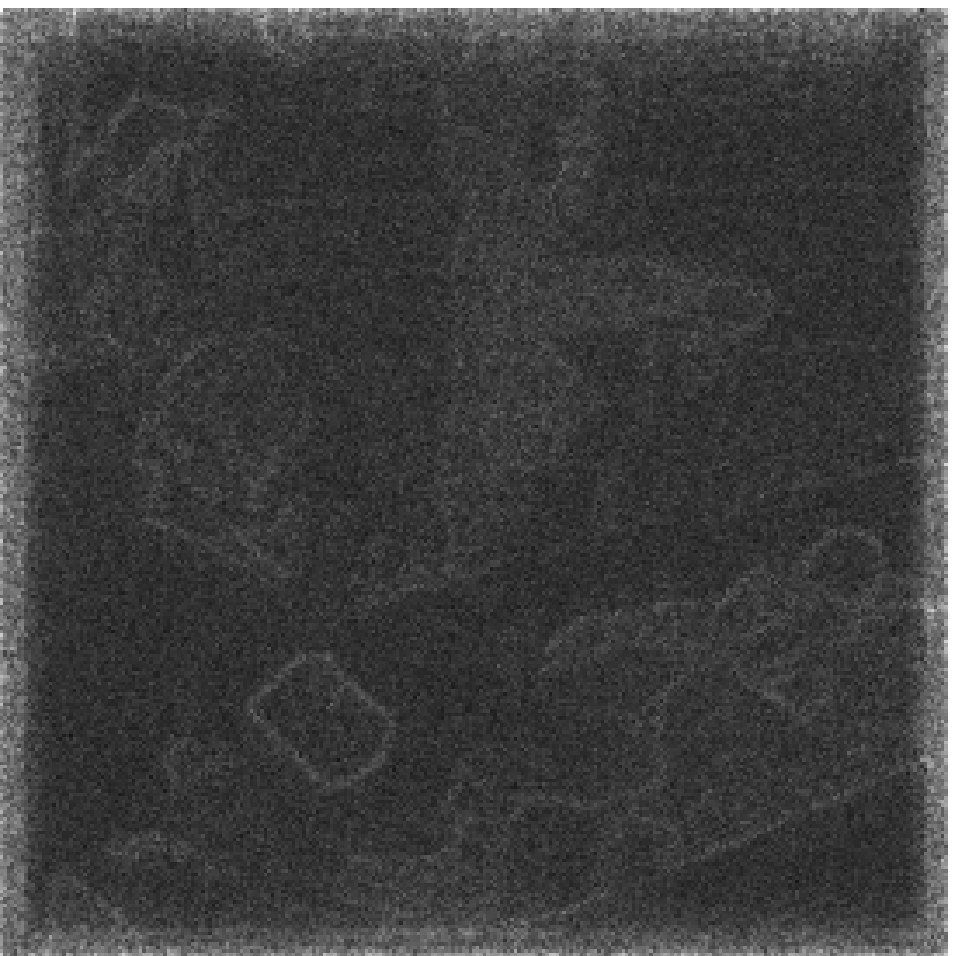}&
    \includegraphics[width=0.45\columnwidth]{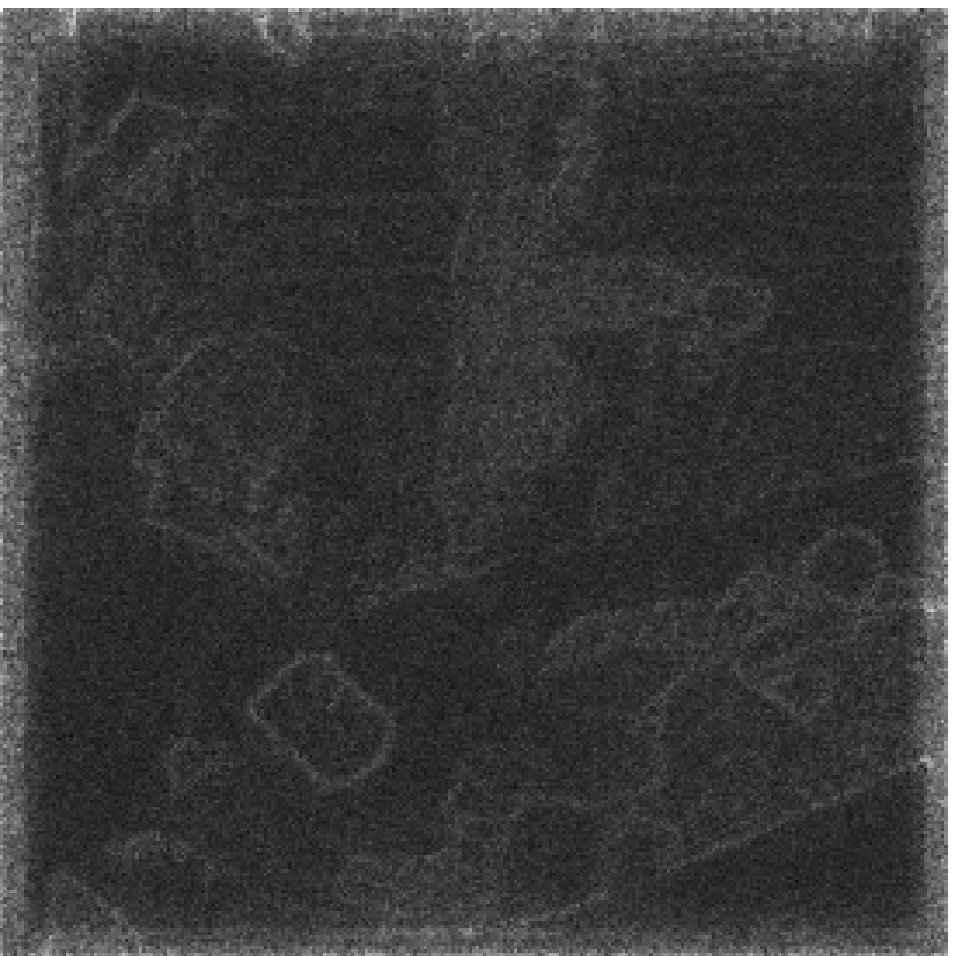}&
    \raisebox{0.2\columnwidth}{\begin{minipage}[c]{0.45\columnwidth}
        \begin{tabular}{c @{ } c}
          \includegraphics[width=0.4\columnwidth]{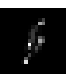}&
          \includegraphics[width=0.4\columnwidth]{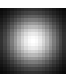}\\
          \includegraphics[width=0.4\columnwidth]{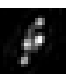}&
          \includegraphics[width=0.4\columnwidth]{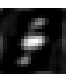}
        \end{tabular}
      \end{minipage}}\\
    (b) blurred {\footnotesize (PSNR=22.57dB)}& (d) VB stdev & (f) EP stdev & (h) kernel evolution\\
  \end{tabular}
  \caption{Image deblurring experiment with the proposed algorithms. (a) Sharp
    \by{255}{255} image. (b) Real blurred image. Posterior mean and pointwise
    estimation uncertainty for non-blind image deblurring under the
    variational bounding (c, d) and expectation propagation (e, f)
    criteria. (g) Blind image deblurring with variational bounding. (h)
    clockwise from upper left, ground-truth \by{19}{19} blurring kernel,
    initialization, and estimated kernels after the first and the final tenth
    EM iteration. The model estimates the image values in an extended
    \by{273}{273} domain, but we only take the central \by{255}{255} area into
    account when calculating the PSNR.}
  \label{fig:deblur}
\end{figure*}

We employ the double-loop algorithms described in \sect~\ref{sec:glm} for both
the variational bounding (VB) and expectation propagation (EP). We use 20
samples for variance estimation, and allow 20 PCG iterations for solving each
of the linear systems~\eqref{eq:pm-sample}. We show the deblurred images from
both the VB and EP algorithms in \fig~\ref{fig:deblur} for both the non-blind
and blind scenaria. Note that EP completely breaks down if we use the Lanczos
variance estimator, while it reliably works under our sample-based variance
estimator.

\section{Discussion}

We have shown that marginal variances required by variational Bayesian
algorithms can be effectively estimated using random sampling. This allows
applying variational Bayesian inference to large-scale problems, essentially
at the same cost as point estimation. The proposed variance estimator can be
thought as a stochastic sub-routine in the otherwise deterministic variational
framework.

Interestingly, efficient Perturb-and-MAP random sampling turns out to be a key
component in both the proposed approach to variational inference and recent
MCMC techniques \cite{PMK08,SGR10,PaYu10,SSR10}. Systematically comparing
these two alternative Bayesian inference alternatives in large-scale
applications arises as an interesting topic for future work.

\subsubsection*{Acknowledgments}
This work was supported by the U.S\onedot Office of Naval Research under the
MURI grant N000141010933 and by the Korean Ministry of Education, Science, and
Technology, under the National Research Foundation WCU program R31-10008. We
thank S\onedot~Lefkimmiatis for suggesting the circulant preconditioner of
\sect~\ref{sec:deblur}, M\onedot~Seeger for his feedback on an earlier version
of the paper, and H\onedot~Nickisch for making the \verb+glm-ie+ toolbox
publicly available.

{\small
\bibliographystyle{ieee}
\bibliography{IEEEabrv,biblio_preamble_abrv,biblio_gpapan}
}

\end{document}

%% file: preamble.tex
\usepackage{graphicx}
\usepackage{multirow}
\usepackage{multirow}
\usepackage{cite}
\usepackage{url}
\usepackage{verbatim}

\usepackage{amsmath}
\usepackage{amsthm}
\usepackage{amssymb}
\usepackage{bm} 

\usepackage{pstricks}
\usepackage{pst-plot}
\usepackage[pstadd]{pst-sigsys}
\usepackage{pstricks-add}

\usepackage[marginclue,inline,final]{fixme}

\newcommand{\Real}{\mathbb{R}} 

\newcommand{\abs}[1]{\ensuremath{\lvert#1\rvert}}
\newcommand{\norm}[1]{\ensuremath{\lVert#1\rVert}}
\newcommand{\Oh}{\mathcal{O}}

\newcommand{\eat}{\ensuremath{\Big\vert_}}

\newcommand{\lspan}[1]{\ensuremath{\mathrm{span}}}
\newcommand{\by}[2]{\ensuremath{#1 \! \times \! #2}}

\newcommand{\trace}{\ensuremath{\mathop{\mathrm{tr}}}}
\newcommand{\diag}{\ensuremath{\mathop{\mathrm{diag}}}}

\newcommand{\argmin}{\ensuremath{\mathop{\mathrm{argmin}}}}
\newcommand{\argmax}{\ensuremath{\mathop{\mathrm{argmax}}}}
\newcommand{\Var}{\ensuremath{\mathop{\mathrm{Var}}}}

\newcommand{\normals}[2]{\ensuremath{\mathcal{N}(#1,#2)}}
\newcommand{\normal}[3]{\ensuremath{\mathcal{N}(#1;#2,#3)}}

\newcommand{\mcomma}{\ensuremath{\,,}}
\newcommand{\mdot}{\ensuremath{\,.}}

\newcommand{\xs}{\ensuremath{x}}        %

\newcommand{\bv}{\ensuremath{\mathbf{b}}}        %
\newcommand{\cv}{\ensuremath{\mathbf{c}}}        %
\newcommand{\gv}{\ensuremath{\mathbf{g}}}        %
\newcommand{\hv}{\ensuremath{\mathbf{h}}}        %
\newcommand{\kv}{\ensuremath{\mathbf{k}}}        %
\newcommand{\rv}{\ensuremath{\mathbf{r}}}        %
\newcommand{\sv}{\ensuremath{\mathbf{s}}}        %
\newcommand{\betav}{\ensuremath{\bm{\beta}}}     %
\newcommand{\gammav}{\ensuremath{\bm{\gamma}}}        %
\newcommand{\xiv}{\ensuremath{\bm{\xi}}}     %
\newcommand{\muv}{\ensuremath{\bm{\mu}}}     %
\newcommand{\zv}{\ensuremath{\mathbf{z}}}        %
\newcommand{\yv}{\ensuremath{\mathbf{y}}}        %
\newcommand{\xv}{\ensuremath{\mathbf{\xs}}}        %
\newcommand{\thetav}{\ensuremath{\bm{\theta}}}        %
\newcommand{\zerov}{\ensuremath{\mathbf{0}}}        %

\newcommand{\Am}{\ensuremath{\mathbf{A}}}        %
\newcommand{\Gm}{\ensuremath{\mathbf{G}}}        %
\newcommand{\Hm}{\ensuremath{\mathbf{H}}}        %
\newcommand{\Um}{\ensuremath{\mathbf{I}}}        %
\newcommand{\Pm}{\ensuremath{\mathbf{P}}}        %
\newcommand{\Rm}{\ensuremath{\mathbf{R}}}        %
\newcommand{\Xm}{\ensuremath{\mathbf{X}}}        %
\newcommand{\Unitm}{\ensuremath{\mathbf{I}}}        %
\newcommand{\Gammam}{\ensuremath{\bm{\Gamma}}}        %
\newcommand{\Sigmam}{\ensuremath{\bm{\Sigma}}}        %

\newcommand{\Sigmap}{\Sigmam}

\newcommand{\fig}{Fig.}
\newcommand{\sect}{Sec.}

\newcommand{\eq}{Eq.}

\newcommand{\expect}[1]{\ensuremath{E\left\{#1\right\}}}
\newcommand{\expectt}[2]{\ensuremath{E_{#1}\left\{#2\right\}}}

\makeatletter 
\DeclareRobustCommand\onedot{\futurelet\@let@token\@onedot}
\def\@onedot{\ifx\@let@token.\else.\null\fi\xspace}
\def\eg{e.g\onedot} 
\def\ie{i.e\onedot} 
\def\cf{c.f\onedot} 
 
\def\vs{vs\onedot} 
\def\wrt{w.r.t\onedot}

\makeatother


%% file: var.bbl
\newcommand{\noopsort}[1]{} \newcommand{\printfirst}[2]{#1}
  \newcommand{\singleletter}[1]{#1} \newcommand{\switchargs}[2]{#2#1}
\begin{thebibliography}{10}\itemsep=-1pt

\bibitem{Atti99}
H.~Attias.
\newblock Independent factor analysis.
\newblock {\em Neur. Comp.}, 11:803--851, 1999.

\bibitem{Bish06}
C.~Bishop.
\newblock {\em Pattern Recognition and Machine Learning}.
\newblock Springer, 2006.

\bibitem{BLN95}
R.~Byrd, P.~Lu, and J.~Nocedal.
\newblock A limited memory algorithm for bound constrained optimization.
\newblock {\em {SIAM} J. Sci. and Statist. Comp.}, 16(5):1190--1208, 1995.

\bibitem{CaTa05}
E.~Candes and T.~Tao.
\newblock Decoding by linear programming.
\newblock {\em {IEEE} Trans. Inf. Theory}, 51(12):4203--4215, Dec. 2005.

\bibitem{CICB10}
V.~Cevher, P.~Indyk, L.~Carin, and R.~Baraniuk.
\newblock Sparse signal recovery and acquisition with graphical models.
\newblock {\em {IEEE} Signal Process. Mag.}, 27(6):92--103, Nov. 2010.

\bibitem{Como94}
P.~Comon.
\newblock Independent component analysis, a new concept?
\newblock {\em Signal Processing}, 36(3):287--314, 1994.

\bibitem{CPT05}
N.~Cressie, O.~Perrin, and C.~Thomas-Agnan.
\newblock Likelihood-based estimation for {G}aussian {MRFs}.
\newblock {\em Stat. Meth.}, 2(1):1--16, 2005.

\bibitem{Dono06}
D.~L. Donoho.
\newblock Compressed sensing.
\newblock {\em {IEEE} Trans. Inf. Theory}, 52(4):1289--1306, Apr. 2006.

\bibitem{FSH+06}
R.~Fergus, B.~Singh, A.~Hertzmann, S.~Roweis, and W.~Freeman.
\newblock Removing camera shake from a single photograph.
\newblock {\em Proc. {SIGGRAPH}}, 25(3):787--794, 2006.

\bibitem{Giro01}
M.~Girolami.
\newblock A variational method for learning sparse and overcomplete
  representations.
\newblock {\em Neur. Comp.}, 13:2517--2532, 2001.

\bibitem{GoVa96}
G.~Golub and C.~Van~Loan.
\newblock {\em Matrix Computations}.
\newblock John Hopkins Press, 1996.

\bibitem{HNL06}
P.~Hansen, J.~Nagy, and D.~O'Leary.
\newblock {\em Deblurring images: matrices, spectra, and filtering}.
\newblock SIAM, 2006.

\bibitem{Jain89}
A.~Jain.
\newblock {\em Fundamentals of digital image processing}.
\newblock Prentice Hall, 1989.

\bibitem{JXC08}
S.~Ji, Y.~Xue, and L.~Carin.
\newblock Bayesian compressive sensing.
\newblock {\em {IEEE} Trans. Signal Process.}, 56(6):2346--2356, June 2008.

\bibitem{JGJS99}
M.~Jordan, J.~Ghahramani, T.~Jaakkola, and L.~Saul.
\newblock An introduction to variational methods for graphical models.
\newblock {\em Machine Learning}, 37:183--233, 1999.

\bibitem{LBU12}
S.~Lefkimmiatis, A.~Bourquard, and M.~Unser.
\newblock Hessian-based norm regularization for image restoration with
  biomedical applications.
\newblock {\em {IEEE} Trans. Image Process.}, 2012.
\newblock to appear.

\bibitem{LWDF09}
A.~Levin, Y.~Weiss, F.~Durand, and W.~Freeman.
\newblock Understanding and evaluating blind deconvolution algorithms.
\newblock In {\em Proc. {CVPR}}, pages 1964--1971, 2009.

\bibitem{LWDF11}
A.~Levin, Y.~Weiss, F.~Durand, and W.~Freeman.
\newblock Efficient marginal likelihood optimization in blind deconvolution.
\newblock In {\em Proc. {CVPR}}, pages 2657--2664, 2011.

\bibitem{LeSe00}
M.~Lewicki and T.~Sejnowski.
\newblock Learning overcomplete representations.
\newblock {\em Neur. Comp.}, 12:337--365, 2000.

\bibitem{Mack92a}
D.~MacKay.
\newblock Bayesian interpolation.
\newblock {\em Neur. Comp.}, 4(3):415--447, 1992.

\bibitem{MJCW08}
D.~Malioutov, J.~Johnson, M.~Choi, and A.~Willsky.
\newblock Low-rank variance approximation in {GMRF} models: Single and
  multiscale approaches.
\newblock {\em {IEEE} Trans. Signal Process.}, 56(10):4621--4634, Oct. 2008.

\bibitem{Mall89}
S.~Mallat.
\newblock A theory for multiresolution signal decomposition: The wavelet
  transform.
\newblock {\em {IEEE} Trans. PAMI}, 11(7):674--693, 1989.

\bibitem{Mall99}
S.~Mallat.
\newblock {\em A Wavelet Tour of Signal Processing}.
\newblock Acad. Press, 2 edition, 1999.

\bibitem{Mink01}
T.~Minka.
\newblock Expectation propagation for approximate bayesian inference.
\newblock In {\em Proc. {UAI}}, 2001.

\bibitem{Nick10}
H.~Nickisch.
\newblock The generalised linear models inference and estimation toolbox
  (glm-ie v. 1.3).
\newblock \url{http://mloss.org/software/view/269}.

\bibitem{OlFi96}
B.~Olshausen and D.~Field.
\newblock Emergence of simple-cell receptive field properties by learning a
  sparse code for natural images.
\newblock {\em Nature}, 381:607--609, 1996.

\bibitem{PaSa82}
C.~Paige and M.~Saunders.
\newblock {LSQR}: An algorithm for sparse linear equations and sparse least
  squares.
\newblock {\em {ACM} Trans. on Math. Soft.}, 8(1):43--71, 1982.

\bibitem{PWKR05}
J.~Palmer, D.~Wipf, K.~Kreutz-Delgado, and B.~Rao.
\newblock Variational {EM} algorithms for non-gaussian latent variable models.
\newblock In {\em Proc. {NIPS}}, 2005.

\bibitem{PMK08}
G.~Papandreou, P.~Maragos, and A.~Kokaram.
\newblock Image inpainting with a wavelet domain hidden {M}arkov tree model.
\newblock In {\em Proc. {ICASSP}}, pages 773--776, 2008.

\bibitem{PaYu10}
G.~Papandreou and A.~Yuille.
\newblock Gaussian sampling by local perturbations.
\newblock In {\em Proc. {NIPS}}, 2010.

\bibitem{ROF92}
L.~Rudin, S.~Osher, and E.~Fatemi.
\newblock Nonlinear total variation based noise removal algorithms.
\newblock {\em Physica D}, 60:259--268, 1992.

\bibitem{SGR10}
U.~Schmidt, Q.~Gao, and S.~Roth.
\newblock A generative perspective on {MRF}s in low-level vision.
\newblock In {\em Proc. {CVPR}}, 2010.

\bibitem{SSR10}
U.~Schmidt, K.~Schelten, and S.~Roth.
\newblock Bayesian deblurring with integrated noise estimation.
\newblock In {\em Proc. {CVPR}}, pages 2625--2632, 2011.

\bibitem{ScWi01}
M.~Schneider and A.~Willsky.
\newblock Krylov subspace estimation.
\newblock {\em {SIAM} J. Sci. Comp.}, 22(5):1840--1864, 2001.

\bibitem{SeNi11b}
M.~Seeger and H.~Nickisch.
\newblock Fast convergent algorithms for expectation propagation approximate
  bayesian inference.
\newblock In {\em Proc. {AISTATS}}, 2011.

\bibitem{SeNi11}
M.~Seeger and H.~Nickisch.
\newblock Large scale bayesian inference and experimental design for sparse
  linear models.
\newblock {\em {SIAM} J. Imaging Sci.}, 4(1):166--199, 2011.

\bibitem{SNPS08}
M.~Seeger, H.~Nickisch, R.~Pohmann, and B.~Sch\"{o}lkopf.
\newblock Bayesian experimental design of magnetic resonance imaging sequences.
\newblock In {\em Proc. {NIPS}}, pages 1441--1448, 2008.

\bibitem{Simo05}
E.~P. Simoncelli.
\newblock Statistical modeling of photographic images.
\newblock In A.~Bovik, editor, {\em Handbook of Video and Image Processing},
  chapter 4.7. Academic Press, 2 edition, 2005.

\bibitem{Tipp01}
M.~Tipping.
\newblock Sparse {B}ayesian learning and the relevance vector machine.
\newblock {\em J. of Mach. Learn. Res.}, 1:211--244, 2001.

\bibitem{GCLH10}
M.~van Gerven, B.~Cseke, F.~de~Lange, and T.~Heskes.
\newblock Efficient {B}ayesian multivariate {fMRI} analysis using a sparsifying
  spatio-temporal prior.
\newblock {\em NeuroImage}, 50:150--161, 2010.

\bibitem{WaJo08}
M.~Wainwright and M.~Jordan.
\newblock Graphical models, exponential families, and variational inference.
\newblock {\em Found. and Trends in Machine Learning}, 1(1-2):1--305, 2008.

\bibitem{WeFr01}
Y.~Weiss and W.~Freeman.
\newblock Correctness of belief propagation in {G}aussian graphical models of
  arbitrary topology.
\newblock {\em Neur. Comp.}, 13(10):2173--2200, 2001.

\end{thebibliography}
